\documentclass[letterpaper, 10 pt, conference]{ieeeconf}  
\IEEEoverridecommandlockouts                              
\usepackage{tabularx}
\usepackage{wrapfig}
\usepackage{xcolor}
\usepackage[pagebackref=false,breaklinks=true,letterpaper=true,colorlinks,bookmarks=false]{hyperref}
\usepackage{tabulary,multirow,multicol,overpic}
\usepackage[ruled,vlined]{algorithm2e}
\usepackage{algorithmic}
\usepackage{tikz}
\usepackage{textcomp}
\usepackage{amsfonts}       
\usepackage{nicefrac}       
\usepackage{microtype}      
\usepackage{graphicx} 
\usepackage{url}
\usepackage{amsmath, amssymb}
\usepackage{bm}
\usepackage{tabularx}
\usepackage{booktabs}
\usepackage{float}
\usepackage{graphicx}
\usepackage[font=small]{caption}
\usepackage{cite}
\newcommand{\E}{\mathbb{E}}

\title{\LARGE \bf
\textbf{Cross Domain Policy Transfer with Effect Cycle-Consistency}
}

\author{Ruiqi Zhu$^{1}$, Tianhong Dai$^{2}$, Oya Celiktutan$^{1}$ 
\thanks{$^{1}$ R. Zhu and O. Celiktutan are with the Department of Engineering, King's College London.}
\thanks{$^{2}$ T. Dai is with the Department of Computing Science, University of Aberdeen.}
}

\begin{document}

\maketitle
\thispagestyle{empty}
\pagestyle{empty}


\begin{abstract}

Training a robotic policy from scratch using deep reinforcement learning methods can be prohibitively expensive due to sample inefficiency. To address this challenge, transferring policies trained in the source domain to the target domain becomes an attractive paradigm. Previous research has typically focused on domains with similar state and action spaces but differing in other aspects. In this paper, our primary focus lies in domains with different state and action spaces, which has broader practical implications, i.e. transfer the policy from robot A to robot B. Unlike prior methods that rely on paired data, we propose a novel approach for learning the mapping functions between state and action spaces across domains using unpaired data. We propose \emph{effect cycle-consistency}, which aligns the effects of transitions across two domains through a \emph{symmetrical optimization} structure for learning these mapping functions. Once the mapping functions are learned, we can seamlessly transfer the policy from the source domain to the target domain. Our approach has been tested on three locomotion tasks and two robotic manipulation tasks. The empirical results demonstrate that our method can reduce alignment errors significantly and achieve better performance compared to the state-of-the-art method. Project page: \url{https://ricky-zhu.github.io/effect_cycle_consistency}.

\end{abstract}

\section{Introduction}

Deep reinforcement learning (DRL) has demonstrated impressive performance in sequential decision-making problems, such as video games \cite{mnih2015human,silver2016mastering}, robotics manipulation \cite{zhu2022deep,zhu2023learning}, and autonomous driving \cite{kiran2021deep}. However, due to the low sample efficiency, training DRL models from scratch has typically limited their application in real-world scenarios where data collection is costly \cite{yu2018towards, zhang2022human}. Consequently, transferring policies trained in the source domain to the target domains has emerged as a promising research direction. Previous works usually focus on the domains with similar state and action spaces but differing in other aspects, such as dynamics \cite{hanna2017grounded, desai2020imitation}. In this paper, we aim to transfer the policy across domains with different state and action spaces, which has a broader applicability, i.e., transferring the pre-trained policy on robot A to a variety of robots with different state spaces and action spaces.

To enable policy transfer across domains with different state spaces and action spaces, a line of prior works has explored the use of graph neural networks (GNN) \cite{zhou2020graph} to address the differences in state and action spaces, leveraging GNN's ability to process graphs of arbitrary sizes. However, these studies often assume that agents possess limbs, each equipped with proprioceptive sensors, to model structural policies with GNN using hand-designed descriptions \cite{wang2017nervenet,huang2020one,kurin2020my}. Another line of prior works focuses on learning invariant representations across agents \cite{gupta2017learning,sermanet2018time}. By training on the latent representation space, policies become invariant to differences in state spaces and action spaces. Nevertheless, these works often hinge on paired trajectories, acquired from pre-trained policies or human labelling, which can be prohibitively expensive to obtain in real-world applications. 
Furthermore, enforcing invariance may not be universally applicable as various types of invariance may offer advantages for diverse downstream tasks, as demonstrated by recent research in the field of self-supervised visual representation learning \cite{tian2020makes}.

\begin{figure}[!t]
	\centering
 \vspace*{0.1cm}
	\includegraphics[width=\columnwidth]{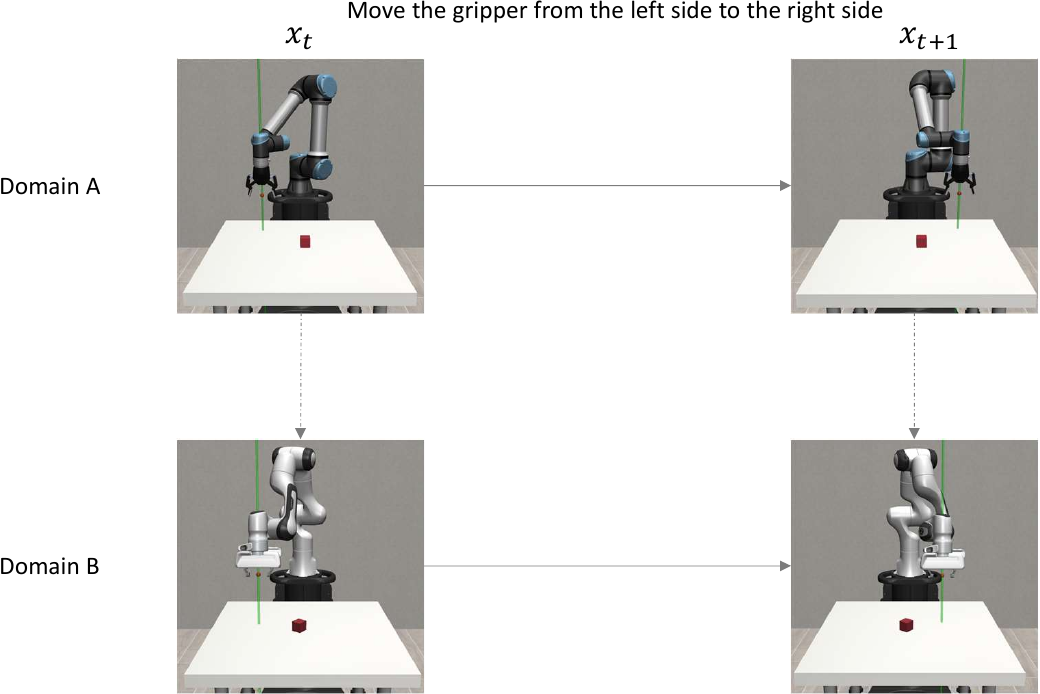}
	\caption{Illustration of the effect cycle-consistency. We aim to align the effect of the transitions across domains. For instance, the effect of transition in Domain A is moving the gripper from the left side to the right side. The effect of the translated transition in the Domain B is expected to move the gripper from the left side to the right side as well. }
	\label{fig: illustration}
  \vspace{-0.5cm}
\end{figure}

Recently, several studies have sought to remove the need for paired data by incorporating the concept of cycle-consistency \cite{zhu2017unpaired} into the process of learning mapping functions of the state spaces and action spaces across domains \cite{rao2020rl, ho2021retinagan}. These investigations have focused on establishing mappings between visual scenes, often overlooking the temporal ordering which indicates the dependency of the consecutive states and actions. In an attempt to incorporate the temporal ordering into the mapping functions learning process, Zhang et al. introduced dynamics cycle-consistency constraint \cite{zhang2020learning}. The constraint enforces that the predicted next state by the forward dynamics model corresponds to the translated next state. However, due to inherent approximation errors associated with the mapping functions and learned forward dynamics models, this method is susceptible to compounding errors. As the time step increases, alignment errors accumulate, leading to a degradation in the performance of the transferred policies. To mitigate the compounding error, Wang et al. proposed to apply weak supervision on the learning of mapping functions using the data with paired abstractions \cite{wang2022weakly}. For example, for ant robots with different legs, the data with paired abstractions refers to the states with the same 2D coordinates but may differ in other dimensions. However, collecting such data is still not trivial and the assumption that the domains have the same abstractions may be violated in certain tasks.

In this paper, we propose a novel framework for learning the mapping functions across the domains with different state spaces and action spaces leveraging unpaired data. Unlike dynamics cycle-consistency which aligns the next states in both domains, we propose \emph{effect cycle-consistenct} that aligns the effect of transitions in both domains to learn the mapping functions as illustrated in Fig. \ref{fig: illustration}. Additionally, we utilize \emph{symmetrical optimization} structure which applies identical objectives to the mapping functions from the source domain to the target domain and vice versa. In summary, the contributions of this paper are:

\begin{itemize}
  \item We propose a novel framework for learning the mapping functions across domains with different state and action spaces.
  \item We propose \emph{effect cycle-consitency} to learn the mapping functions with \emph{symmetrical optimization} structures.
  \item We have conducted experiments on 3 locomotion tasks and 2 robotic manipulation tasks. The empirical results demonstrate that our method achieves better performance with the transferred policies and lower alignment errors.
\end{itemize}

\begin{figure*}[!th]
	\centering
        \vspace*{0.1cm}
	\includegraphics[width=2\columnwidth]{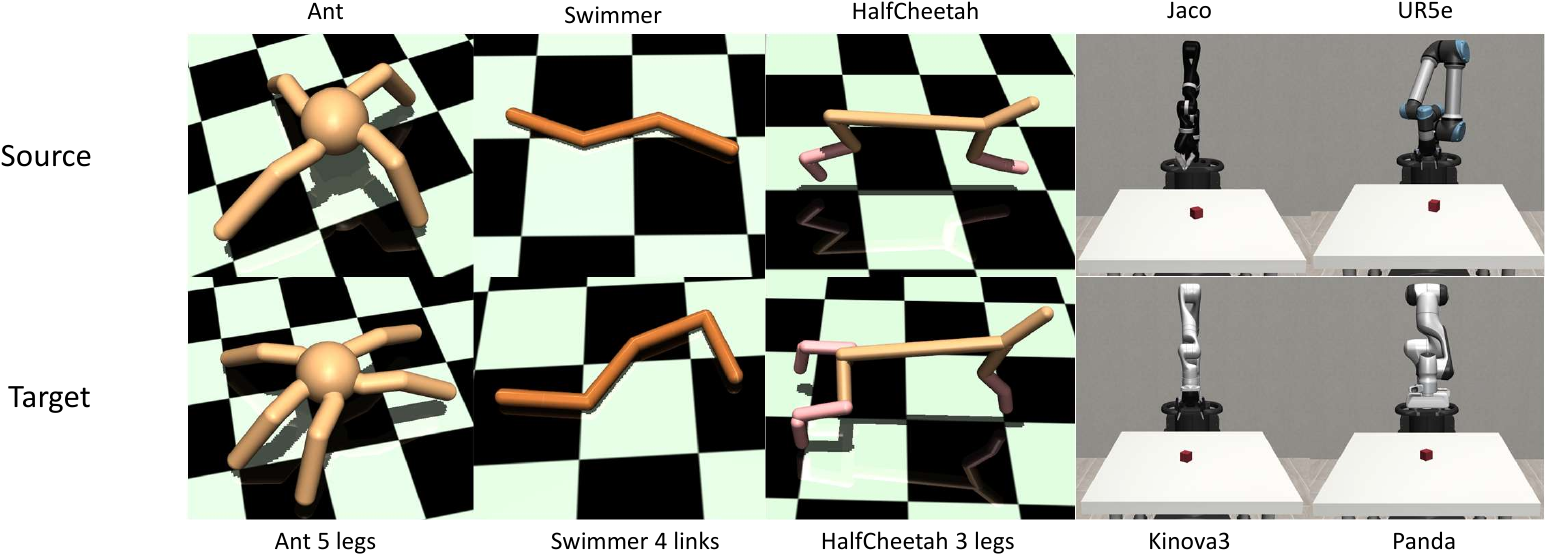}
        \caption{Visulization of the agents in the source domains and the target domains.}
	\label{fig: demo_morph}
 \vspace{-0.5cm}
\end{figure*}

\section{Related Works}
\noindent\textbf{Cross Morphology Policy Transfer.} 
To transfer a pre-trained policy across agents with different morphologies, a line of prior works model the agents using GNN for leveraging the capability of GNN to process graphs of arbitrary sizes \cite{wang2017nervenet, kurin2020my}. These works usually assume that the agents have rigid structures such as limbs. Additionally, they need hand-designed descriptions to model the graphs, which is non-trivial. To remove the need for the hand-designed description, Trabucco et al. propose a data-driven method which trains a transferrable policy among a broad set of morphologies \cite{trabucco2022anymorph}. The policy in the approach is conditioned on the agent's state and action, which are used as inputs to a neural network that infers the agent's morphology. The inferred morphology is then used to train a policy that can generalize to new morphologies. However, the approach requires the careful design of the training set of morphologies to ensure the testing morphology representations lie within the support of the distribution. Hejna et al. proposed an approach for transferring the policy by leveraging the similarity of the morphologies \cite{hejna2020hierarchically}. The approach aligns the pre-trained low-level policies and then learns transferrable high-level policies. However, the approach is limited to transfer policies between similar morphologies.\\

\noindent\textbf{Learning Invariant Representation.}  To learn the mapping across domains, researchers have proposed to learn the representations which are invariant to the factors that are irrelevant to the downstream tasks and only preserve task-specific information. 
Domain randomization methods learn invariant task-specific representation by randomizing some factors of the domains, such as lights, and textures \cite{peng2018sim, tobin2017domain,andrychowicz2020learning}. The policies trained in the augmented domains are supposed to be robust to the variations of these factors. Nevertheless, these methods require expertise to determine what factors to randomize. Also, they assume the distribution of these factors of the target domains lies within the support of the randomization distribution, which could be violated in practical applications. Another line of prior work instead attempts to discover the invariant representation with paired trajectories of the source domains and the target domains \cite{gupta2017learning, yan2021learning,liu2018imitation}. However, collecting the paired trajectories usually requires human annotations, which could be expensive and tedious \cite{taylor2009transfer}. In contrast to these methods, we aim to discover the mapping across the source domains and target domains using unpaired data, which would be more applicable.\\

\noindent\textbf{Cycle-Consistency.} To remove the need for paired data in the field of image-to-image translation, Zhu et al. proposed cycle-consistency \cite{zhu2017unpaired} to discover the correspondence between different image domains leveraging Generative Adversarial Networks \cite{goodfellow2020generative}. Subsequently, the method has been extended to other fields, such as unsupervised video retargeting \cite{bansal2018recycle}, and domain adaptation \cite{hoffman2018cycada}. Recently, the concept of cycle-consistency has been brought to the field of sim-to-real adaptation for robotic tasks \cite{rao2020rl,ho2021retinagan}. However, these applications typically are limited to visual adaptations. Zhang et al. have proposed dynamics cycle-consistency to
learn the mapping functions across the state spaces and the target spaces of different domains \cite{zhang2020learning}. The approach has integrated the dynamics information into learning the mapping functions and hence can distinguish the temporal ordering. The approach achieves state-of-the-art performance in policy transfer across domains with different state spaces and action spaces using unpaired data.

\section{Background}
\noindent\textbf{Problem Setting.} 
The source domain and target domain are both modelled as Markov Decision Process (MDP). We model the source domain as $\mathcal{M}^1=\{X, A,\mathcal{T}^{1},\mathcal{R}^{1},p_{0}^{1},\gamma\}$ and the target domain as $\mathcal{M}^2=\{Y, U,\mathcal{T}^{2},\mathcal{R}^{2},p_{0}^{2},\gamma\}$. The source domains and the target domains have different state spaces and action spaces. We aim to transfer the pre-trained policy in the source domain to the target domain for solving the same tasks. 

Owing to the mismatch of the state spaces and the action spaces, it is essential to define the mapping functions across the domains. Specifically, we denote the state mapping function and action mapping function from the source domain to the target domain as $F:X\rightarrow Y$ and $H:X \times A\rightarrow U$ respectively. Similarly, we denote the state mapping function and action mapping function from the target domain to the source domain as $G:Y\rightarrow X$ and $P:Y \times U\rightarrow A$ respectively.

We aim to utilize unpaired data to learn the mapping functions.
We formulate the dataset of the source domain as $\tau_{\mathcal{M}^1}=\{x_{t},a_{t},x_{t+1}\}_{i}$ and the dataset of the target domain as $\tau_{\mathcal{M}^2}=\{y_{t},u_{t},y_{t+1}\}_{i}$. The unpaired datasets contain no task-specific behaviours. The mapping functions can be used to transfer the pre-trained policy to the target domain as shown in Algo. \ref{algo:deploy}.

\begin{figure*}[!t]
	\centering
 \vspace*{0.1cm}
	\includegraphics[width=2\columnwidth]{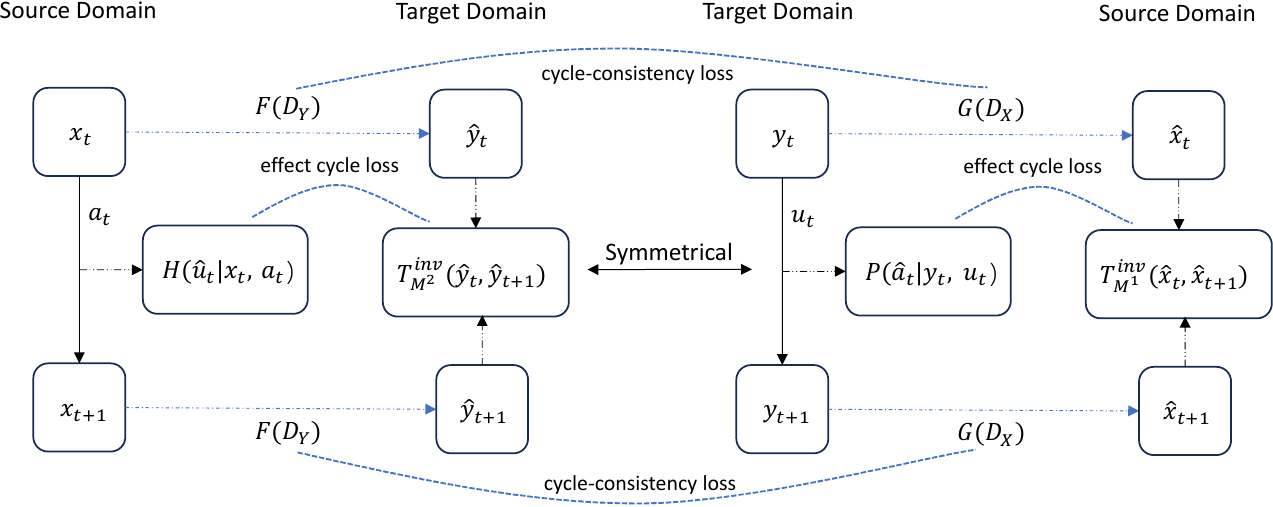}
        \caption{Illustration of our method.}
	\label{fig: network}
 \vspace{-0.5cm}
\end{figure*}

\noindent\textbf{Dynamics cycle consistency.} To discover the mapping functions across the domains, Zhang et al. proposed dynamics cycle-consistency which integrates dynamics information into the learning while leveraging adversarial objectives \cite{zhang2020learning}. It learns the state mapping function from the target domain to the source domain using adversarial objectives as:
\begin{equation}
\begin{aligned}
    \min_{G}\max_{D_{X}}  \ & \E_{x\sim \tau_{\mathcal{M}^1}}[\log D_{X}(x)] + \\
    &\E_{y\sim \tau_{\mathcal{M}^2}}[\log (1-D_{X}(G(y)))] 
\end{aligned}
\end{equation}
where $D_{X}$ is the discriminator which tries to distinguish the translated sample $G(y)$ against real samples $x$. Similarly, it utilizes the adversarial objectives to learn the action mapping functions $H$ and $P$. Additionally, it utilizes cycle-consistency constraints on learning the action mapping functions as:
\begin{equation}
\label{eq:dyn_clc_1}
\begin{aligned}
    &\min_{H,P} \E_{a\sim \tau_{\mathcal{M}^1 }}[\lVert P(y,H(x,a))-a \rVert_{1}]
\end{aligned}
\end{equation}
Following that, the approach integrates the dynamics information by imposing the translated next state corresponds to the predicted next state by the forward dynamics model $T_{\mathcal{M}_{2}}$ as:
\begin{equation}
\label{eq:dyn_clc_2}
\begin{aligned}
    \min_{F,H} \E_{(x_{t},a_{t},x_{t+1})\sim \tau_{\mathcal{M}^1}}[\lVert &F(x_{t+1})-\\
    &T_{\mathcal{M}_{2}}(F(x_{t}),H(x_{t},a_{t}))\rVert_{1}]
\end{aligned}
\end{equation}
With this objective, the learned mapping function can be aware of the temporal ordering, which would be ignored if simply applying cycle-consistency loss on learning the mapping functions. However, the approach is susceptible to compounding errors. For instance, given a trajectory in the source domain $\tau = \{x_{0},a_{0},...,s_{T}\}$, two methods to derive the translated state at time step $T$ are : (1) $y_{T}=F(x_{T})$; (2) $\hat{y}_{T}=T_{\mathcal{M}_{2}}(\cdots T_{\mathcal{M}_{2}}(F(x_{0}),H(x_{0},a_{0})),...,H(x_{T},a_{T}))$. The second method progressively utilizes the forward dynamics model and the mapping functions to generate the next states. Hence, the approximation errors associated with the forward dynamics model and the mapping functions will enlarge the alignment errors between the translated states using the two methods with longer horizons \cite{wang2022weakly}. Consequently, the large alignment errors could potentially degrade the performance of the transferred policies.

\section{Effect Cycle-Consistency with Symmetrical Optimization}
In this section, we present our method for solving policy transfer across domains with different state and action spaces using unpaired data. Our framework is illustrated in Fig. \ref{fig: network}.

Our initial step involves learning state mapping functions with adversarial training objectives, as defined in Eq. \ref{eq:1} and Eq. \ref{eq:2}. Given unpaired samples $x_{i}\in X$ and $y_{i}\in Y$ from the dataset, the state mapping functions $G(y)$ and $F(x)$ aim to map the state distribution to that of their counterparts, while the discriminators $D_{X}$ and $D_{Y}$ aim to distinguish between the real samples and the translated samples \cite{goodfellow2020generative}. Additionally, to prevent mode collapse, we include a cycle-consistency loss \cite{zhu2017unpaired}, as defined in Eq. \ref{eq:3}, in the training process. Consequently, the translated sample is expected to correspond to the original sample when translated back.

\begin{equation}
\begin{aligned}
    \min_{G}\max_{D_{X}} \ &\mathcal{L}_{adv}(G,D_{X})=\E_{x\sim \tau_{\mathcal{M}^1 }}[\log D_{X}(x)] +\\ &\E_{y\sim \tau_{\mathcal{M}^2 }}[\log (1-D_{X}(G(y)))] \label{eq:1}
\end{aligned}
\end{equation}
\begin{equation}
\begin{aligned}
    \min_{F}\max_{D_{Y}} \ &\mathcal{L}_{adv}(F,D_{Y})=\E_{y\sim \tau_{\mathcal{M}^2}}[\log D_{Y}(y)] + \\ &\E_{x\sim \tau_{\mathcal{M}^1}}[\log (1-D_{Y}(F(x)))] \label{eq:2}
\end{aligned}
\end{equation}
\begin{equation}
\begin{aligned}
    \min_{G,F} \ &\mathcal{L}_{cyc}(G,F)=\E_{x\sim \tau_{\mathcal{M}^1}}[\lVert x - G(F(x)) \rVert_{1}]+\\ &\E_{y\sim \tau_{\mathcal{M}^2}}[\lVert y - F(G(y)) \rVert_{1}] \label{eq:3}
\end{aligned}
\end{equation}


For the adversarial training objectives, it achieves the global optimal when the probabilities given by the discriminators equal the ratio of the real samples. For the cycle-consistency objective, it achieves the global optimal when the state mapping functions successfully establish one-to-one mappings between domains. However, despite achieving the global optimum for both objectives, misalignment issues persist due to the inability to learn the temporal ordering \cite{zhang2020learning}. For instance, given transition tuples $(x_{t},a_{t}, x_{t+1})$ and $(y_{t},u_{t}, y_{t+1})$, $G$ can map $x_{t}$ to $y_{t+1}$ and $F$ can still map back $y_{t+1}$ to $x_{t}$, which does not the violate cycle-consitency constraint. However, $x_{t}$ and $x_{t+1}$ are supposed to be mapped to $y_{t}$ and $y_{t+1}$ respectively.

To enable the mapping functions to be aware of the temporal ordering, we incorporate dynamics information into the learning by proposing \emph{effect cycle-consistency} constraints which align the effect of the transitions between domains as shown in Eq. \ref{eq:act_mapping_train}. $T_{\mathcal{M}^1}^{inv}$ and $T_{\mathcal{M}^2}^{inv}$ represent the inverse dynamics models of the source domain and the target domain respectively. They predict the actions that lead to a given transition. Given the tuple $(x_{t},a_{t},x_{t+1})$ from the source domain, the translated action distribution $H(\hat{u}_{t}|x_{t},a_{t})$ is supposed to correspond to the predicted target action distribution given by the inverse dynamic model of the target domain $T_{\mathcal{M}^2}^{inv}(u_{t}|F(x_{t}),F(x_{t+1}))$ with the translated consecutive states as inputs. 

\setlength{\abovedisplayskip}{3pt} 
\setlength{\belowdisplayskip}{6pt} 
\begin{equation}
\label{eq:act_mapping_train}
\begin{aligned}
    &\min \mathcal{L}_{eff}(F,H)=D_{KL}(T_{\mathcal{M}^2}^{inv}(F(x_{t}),F(x_{t+1}))|| H(x_{t},a_{t}))\\
    &\min \mathcal{L}_{eff}(G,P)=D_{KL}(T_{\mathcal{M}^1}^{inv}(G(y_{t}),G(y_{t+1}))||P(y_{t},u_{t}))
\end{aligned}
\end{equation}

However, as the inverse dynamics model represents the physical property of the domain, it is not differentiable for back-propagation. Thus, we train neural networks to approximate the inverse dynamics models of both domains $T_{\mathcal{M}^1}^{inv}$ and $T_{\mathcal{M}^2}^{inv}$. We utilize a supervised objective to train the inverse dynamics models using the dataset to achieve that. As shown in Eq. \ref{eq:inverse_train1} and Eq. \ref{eq:inverse_train2}, to reduce the variance during the training, we leverage the reparameterization trick \cite{kingma2013auto}. We afterwards fix the parameters of the inverse dynamics models throughout the training of the mapping functions.

\begin{equation}
\label{eq:inverse_train1}
\begin{aligned}
   &\min \mathbb{E}_{\substack{(x_{t},a_{t},x_{t+1})\sim \tau_{\mathcal{M}^1}\\\epsilon \sim \mathcal{N}(0,I)}} [\lVert a_{t}-(\mu+\epsilon*\sigma)\rVert_{1}] \\
    s.t.\ \ &\mu = \text{\emph{mean}\ ($T_{\mathcal{M}^1}^{inv}(x_{t},x_{t+1})$)} \\
    & \sigma = \text{\emph{std}\ ($T_{\mathcal{M}^1}^{inv}(x_{t},x_{t+1})$)}
\end{aligned}
\end{equation}

\begin{equation}
\label{eq:inverse_train2}
\begin{aligned}
   &\min \mathbb{E}_{\substack{(y_{t},u_{t},y_{t+1})\sim \tau_{\mathcal{M}^2}\\\epsilon \sim \mathcal{N}(0,I)}} [\lVert u_{t}-(\mu+\epsilon*\sigma)\rVert_{1}] \\
    s.t.\ \ &\mu = \text{\emph{mean}\ ($T_{\mathcal{M}^2}^{inv}(y_{t},y_{t+1})$)} \\
    & \sigma = \text{\emph{std}\ ($T_{\mathcal{M}^2}^{inv}(y_{t},y_{t+1})$)}
\end{aligned}
\end{equation}

Both the outputs of the action mapping functions and the inverse dynamics models are formulated as Gaussian distributions. We minimize the Kullback-Leibler divergences between the two distributions. Hence, we can have analytic solutions to the objectives. Additionally, using $D_{KL}(T_{\mathcal{M}^2}^{inv}||H)$ instead of $D_{KL}(H||T_{\mathcal{M}^2}^{inv})$ ensures that the learned action mapping function $H$ is mode-covering \cite{bishop2006pattern}, i.e., represents all actions captured by the inverse dynamics model.

 As $H(u_{t}|x_{t},a_{t})=\sum T(x_{t+1}|x_{t},a_{t})H(u_{t}|x_{t},a_{t},x_{t+1})$ where $T$ is the characteristic transition model of the source domain, therefore the objective can be interpreted as enabling the action mapping function to predict the action which can lead to the translated transition $(F(x_{t}),F(x_{t+1}))$ given the transition $(x_{t},x_{t+1})$ conditioned on the action $a_{t}$. In contrast to dynamics cycle-consistency which is susceptible to compounding errors caused by the progressive reliance on single-step prediction of the next state, we instead align the effect of the transitions, which is independent of the progressivity. By removing the dependence on the progressivity, the compounding errors will be mitigated as the errors only come from the alignment errors of the mapping functions and will not compound with time steps increase. Additionally, the temporal ordering is embedded as the action distributions given by $T_{\mathcal{M}^2}^{inv}(F(x_{t}),F(x_{t+1}))$ and $T_{\mathcal{M}^2}^{inv}(F(x_{t+1}),F(x_{t}))$ are different. As the effect cycle-consistency loss also backpropagates through the state mapping functions, the learned state mapping functions should be aware of the temporal ordering information.

Note in the training of the mapping functions, the same learning objectives are applied to the mapping functions from the source domain to the target domain and the mapping functions from the target domain to the source domain as shown in Fig. \ref{fig: network}. We refer to the optimization structure as \emph{symmetrical optimization} structure. With the optimization structure, the learned mapping functions in the source(target) domains will influence those in the target(source) domains through the cycle-consistency loss. The optimization structure has demonstrated its significance in improving the performance of the transferred policies and stabilizing the training as shown in Section \ref{sec:exp}. Overall. the full objective is given as:

\begin{equation}
\label{eq:full_obj}
\begin{aligned}
   \mathcal{L}_{full} = &\lambda_{1}(\mathcal{L}_{adv}(G,D_{X})+\mathcal{L}_{adv}(F,D_{Y})+\mathcal{L}_{cyc}(G,F)) \\&+ \lambda_{2}(\mathcal{L}_{eff}(F,H)+\mathcal{L}_{eff}(G,P)).
\end{aligned}
\end{equation}

As in the training of the mapping functions, it involves joint optimization of multiple neural networks, directly optimizing full objectives could lead to trivial solutions. Hence, we utilize the alternating training procedure. Specifically, we fix the parameters of the action mapping functions when optimizing adversarial objectives and the cycle-consistency objective by setting $\lambda_{2}$ as $0$. We set $\lambda_{1}$ as $0$ when optimizing the \emph{effect cylce-consistency} objective as shown in Algo. \ref{algo:training}.

\begin{table*}[t]
  \centering
  \caption{The performance of the transferred policy under different morphologies. (w.o. denotes without)}
  \label{tab:performance}
  \scriptsize
\begin{tabular}{llc|ccccc}
\toprule
Source X& Target $Y$ & Oracle, X & Random &Cycle-GAN &DCC & Ours, $X{\rightarrow}Y$ &Ours w.o. symmetrical\\
\midrule
HalfCheetah & HalfCheetah 3 legs & $6773.36\scriptstyle{\pm43.65}$ &$-266.03\scriptstyle{\pm115.24}$&$-65.76\scriptstyle{\pm32.23}$ & $1361.65\scriptstyle{\pm175.96}$ & \textbf{1981.36$\pm$72.81} &$1427.25\scriptstyle{\pm102.42}$ \\
Swimmer & Swimmer 4 links & $301.81\scriptstyle{\pm6.25}$ & $-4.86\scriptstyle{\pm7.53}$&$8.43\scriptstyle{\pm7.97}$ & $53.73\scriptstyle{\pm18.29}$ &\textbf{207.28$\pm$21.87} & $55.09\scriptstyle{\pm32.52}$\\
Ant & Ant 5 legs & $4319.85\scriptstyle{\pm49.39}$ & $-213.55\scriptstyle{\pm195.65}$&$28.97\scriptstyle{\pm52.47}$ & $670.53\scriptstyle{\pm189.89}$ &\textbf{872.28$\pm$77.80} &$720.21\scriptstyle{\pm159.12}$\\
Jaco & Kinova3 & $45.32\scriptstyle{\pm3.74}$ & $0.57\scriptstyle{\pm0.53}$&$1.57\scriptstyle{\pm1.23}$ & $8.53\scriptstyle{\pm7.25}$ &\textbf{11.25$\pm$6.35} &$7.12\scriptstyle{\pm5.13}$\\
UR5e & Panda & $45.99\scriptstyle{\pm1.44}$ & $1.12\scriptstyle{\pm0.34}$&$1.87\scriptstyle{\pm0.73}$ & $6.89\scriptstyle{\pm5.36}$ &\textbf{7.34$\pm$5.15} &$7.01\scriptstyle{\pm5.82}$\\
\bottomrule
\end{tabular}
\end{table*}

\begin{algorithm}[tb] 
    \SetAlgoLined
    \SetAlgoNoEnd
    \caption{Evaluation}
    \label{algo:deploy}
    {\bfseries Input:} state mapping function $G:Y \rightarrow X$, action mapping function $H:X \times A \rightarrow U$, pre-trained policy in the source domain $\pi(a|x)$ \\
    \For{$t=1...H$}{
    Observe $y_{t}$ ;\\
    Translate to the source domain $x_{t}=G(y_{t})$;\\
    Obtain the action in the source domain $a_{t}\sim \pi(a_{t}|x_{t})$;\\
    Translate the action to the target domain $u_{t}\sim H(u_{t}|x_{t},a_{t})$ ;\\
    Execute $u_{t};$
    
    }

\end{algorithm}

\begin{algorithm}[!h] 
    \SetAlgoLined
    \SetAlgoNoEnd
    \caption{Training}
    \label{algo:training}
    {\bfseries Input:} state mapping function $G:Y \rightarrow X$ and $F:X \rightarrow Y$, action mapping function $H:X \times A \rightarrow U$ and $P:Y \times U \rightarrow A$, dataset of the source domain $\tau_{\mathcal{M}^1}$ and the dataset of the target domain $\tau_{\mathcal{M}^2}$. \\
    Train the inverse dynamics models $T_{\mathcal{M}^1}^{inv}$ and $T_{\mathcal{M}^2}^{inv}$ using Eq. \ref{eq:inverse_train1} and Eq. \ref{eq:inverse_train2};\\
    \For{$i=1..e$}{
    reset $\lambda_{1}$, set $\lambda_{2}=0$;\\
    \For{$i=1...e_{1}$}{
    Train $G$ and $F$ using $\mathcal{L}_{full}$ as shown in Eq. \ref{eq:full_obj};\\
    }
    reset $\lambda_{2}$, set $\lambda_{1}=0$;\\
    \For{$i=1...e_{2}$}{
    Train all the mapping functions using $\mathcal{L}_{full}$ as shown in Eq. \ref{eq:full_obj};\\
    }
    }
\end{algorithm}

\section{Experiment} \label{sec:exp}
We aim to answer the following questions through our
experiments: \textbf{(1)} How is the efficacy of our methods compared to the baselines? \textbf{(2)} Does our method reduce the alignment errors compared to the baselines? \textbf{(3)} What is the impact of the dataset size? \textbf{(4)} What is the importance of the \emph{symmetrical optimization} structure in our methods?

\subsection{Experiment Setup}
The pre-trained policies in the source domain are trained with TD3 algorithms \cite{fujimoto2018addressing} for all tasks. The dataset used in the experiment contains $1k$ unpaired trajectories collected by random policies in both domains.
To answer the above questions, we have designed our experiments from $3$ aspects.

\subsubsection{The efficacy of our proposed method}

To compare the proposed method with baselines, we have evaluated the performance of the transferred policy in the target domains through the state mapping functions and action mapping functions as depicted in Algo. \ref{algo:deploy}. 

We have conducted experiments on $3$ locomotion tasks based on Mujoco \cite{todorov2012mujoco} and $2$ robotic reaching tasks based on Robosuite \cite{zhu2020robosuite}. Additionally, we have added 2D coordinates of the agents to the observation spaces for the locomotion tasks. For each task, the source domains and the target domains have different state and action spaces. The specifications of the state and action spaces are shown in Table \ref{tab:morphology_setup}. And the visualization of the agents in the source domains and the target domains are shown in Fig. \ref{fig: demo_morph}.

\textbf{Baselines.} We have compared our methods with: \textbf{(1) Random}, in which the mapping functions across source and target domains are random; \textbf{(2) Cycle-GAN}, which imposes cycle-consistency loss \cite{zhu2017unpaired} on learning the state mapping functions and the action mapping functions respectively; \textbf{(3) DCC \cite{zhang2020learning}}, which utilizes dynamic cycle-consistency in learning the mapping functions. We follow the official implementation of this baseline. \textbf{(4) Ours w.o. symmetrical}, which ablates the symmetrical optimization structure in our method by removing the action mapping function $P$ as it is not used in the evaluation as shown in Algo.\ref{algo:deploy}. 

\subsubsection{Alignment error evaluation}
To evaluate the alignment errors, we focus on two locomotion tasks (HalfCheetah to HalfCheetah 3 legs, Ant to Ant 5 legs) as we do not have access to the ground truth of translated states in the target domain for the evaluation. We assume the 2D coordinates of the agents should remain the same after being translated by the state mapping functions. Thus, We utilize the 1-norm error between the coordinates of the original states and the translated states as a proxy to evaluate the alignment errors.

\subsubsection{The impact of different dataset sizes}
To investigate the impact of dataset size, We have trained the mapping functions with the datasets containing $2k$, $5k$, $10k$ and $20k$ unpaired trajectories respectively for the task (HalfCheetah to HalfCheetah 3 legs). We then evaluate the performance of the transferred policy to demonstrate the impact.

\begin{table}[ht]
    \vspace{-5pt}
    \centering
        \caption{Dimensions of state spaces and action spaces of the source domains and the target domains.}
    \label{tab:morphology_setup}
    \resizebox{.49\textwidth}{!}{
    \begin{tabular}{cc|cc|ccc}
    \toprule
\multirow{2}{50pt}{Source $\mathcal{M}^1$} & \multirow{2}{50pt}{Target $\mathcal{M}^2$} &\multicolumn{2}{c|}{$\mathcal{M}^1$} & \multicolumn{2}{c}{$\mathcal{M}^2$}\\
&&State & Action & State & Action \\
    \midrule
    HalfCheetah & HalfCheetah 3 legs &  19 &  6  &  25 &  9  \\
    Swimmer & Swimmer 4 links &  10 &  2  &  12 &  3 \\
    Ant & Ant 5 legs &  113 &  8  &  115 &  10 \\
    Jaco & Kinova3 &  50 &  7  &  50 &  8 \\
    UR5e & Panda &  47 &  7  &  42 &  8 \\
    \bottomrule
    \end{tabular}}
    \vspace{-5pt}
\end{table}

\subsection{Experiment Results}
\subsubsection{The efficacy of our proposed method}
For each task, we trained our methods and baselines with 5 different seeds and evaluated the performance with 10 episodes. We report the average and the standard deviation of performances in Table \ref{tab:performance}. As shown in the table, since cycle-GAN cannot recover the temporal ordering information in learning the mapping functions even though the state and action distributions of both domains in the dataset can be translated, the performance is poor. Compared to DCC, the state of the arts under this problem setting, our proposed method outperforms it in all the locomotion tasks and robotic reaching tasks. The outperformance ranges from \textbf{6\%} (UR5e to Panda) to \textbf{285.8\%} (Swimmer to Swimmer 4 links). Additionally, the performance of our proposed methods presents lower variance, which reflects the stability of the performance. We have also evaluated the performance of our proposed method without the symmetrical optimization structure (Ours w.o. symmetrical). The results show that the performance has dropped dramatically and the variance has increased, which demonstrates the importance of the optimization structure.

\subsubsection{The alignment error analysis}
\begin{figure}[!h]
	\centering
	\vspace*{0.1cm}
	\includegraphics[width=\columnwidth]{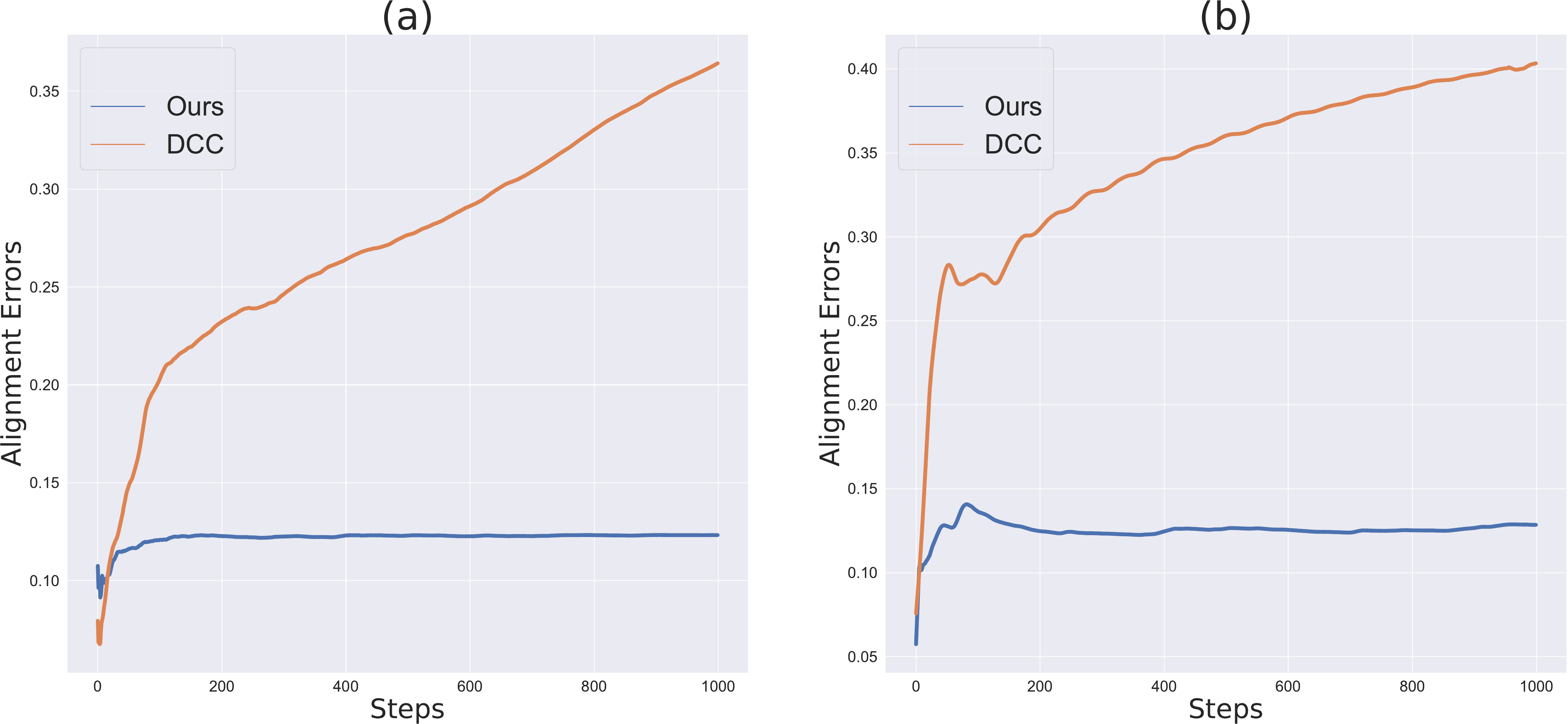}
        \caption{The alignment errors v.s. time steps. (a) HalfCheetah to HalfCheetah 3 legs; (b) Ant to Ant 5 legs.}
        \label{fig:err_analysis}
 \vspace{-0.5cm}
\end{figure}

We have compared our proposed method to DCC with respect to the alignment errors. We rollout the transferred policies in the target domain for 5 different seeds using the trained mapping functions and report the smoothed average of the running averaging of the alignment errors v.s. the time steps. With the time steps increase, the alignment errors of DCC present a trend of increasing which corresponds to the compounding errors as shown in Fig. \ref{fig:err_analysis}. In contrast, our proposed method presents lower alignment errors, and the compounding errors have been significantly reduced.

\subsubsection{The impact of different dataset sizes}
We have trained our proposed method for 5 different seeds using the dataset of different sizes. We report the average and the standard deviation of the performance in Table. \ref{table:diff_scale}. With the size increase, the inverse dynamics models can have better approximations and therefore lead to better action mapping functions and state mapping functions through \emph{effect cycle-consistency} constraints, which further offers a better performance of the transferred policies. However, as the distribution of the dataset does not match the distribution of the encountered data during evaluation, the trained mapping functions cannot generalize to the unseen data in the evaluation. Consequently, as the scale increases from $10k$ to $20k$, the performance does not change much. 

\begin{table}[!h]
    \small
    \vspace{-5pt}
    \centering
        \caption{The performance of using different dataset sizes.}
    \begin{tabular}{c|c}
    \toprule
    Scale & Performance \\
    \midrule
    1k & $1981.36\scriptstyle{\pm72.81}$   \\
    2k & $2061.41\scriptstyle{\pm106.96}$  \\
    5k & $2113.12\scriptstyle{\pm85.67}$  \\
    10k & $2307.63\scriptstyle{\pm101.91}$   \\
    20k & $2282.77\scriptstyle{\pm186.41}$   \\
    \bottomrule
    \end{tabular}
    \vspace{-0.5cm}
    \label{table:diff_scale}
\end{table}

\section{Conclusion}
In this paper, we present a novel framework that leverages unpaired data to learn the mapping functions across domains with different state spaces and action spaces. Within this framework, we propose \emph{effect cycle-consistency} that align the effects of the original transitions and the translated transitions and \emph{symmetrical optimization structure} in learning the mapping functions. The empirical results demonstrate that our proposed method obtains lower alignment errors and better performance compared to the baselines. With this framework, a learned policy can be seamlessly transferred to other agents without requiring task-specific datasets, which greatly alleviates the burden brought by the sample inefficiency of DRL methods. An exciting direction for future work would be generalizing the mapping functions to unseen data in the evaluation to further improve the performance.
\section{Acknowledgement}
This work of Ruiqi Zhu is supported by the King's China Scholarship Council (K-CSC) PhD Scholarship programme.
\newpage

\bibliography{main}
\bibliographystyle{IEEEtran}
\end{document}